\documentclass[11pt,a4paper]{article}

\usepackage{url}            
\usepackage{booktabs}       
\usepackage{amsfonts}       
\usepackage{nicefrac}       
\usepackage{microtype}      
\usepackage{float,subfig}
\usepackage[svgnames]{xcolor}
\usepackage{mathtools}
\usepackage{upgreek}
\usepackage{soul}

\usepackage[utf8]{inputenc}
\usepackage{array, makecell} %

\numberwithin{equation}{section}
\usepackage{hyperref}
\usepackage{amssymb,amsmath,amsthm,enumitem}
\hypersetup{linktoc=all}
\usepackage{physics} 
\usepackage{bibentry}
\usepackage{algorithm}
\usepackage[noend]{algpseudocode}
\usepackage[english]{babel}
\usepackage{blindtext}
\usepackage[export]{adjustbox}
\usepackage{calc}
\usepackage[subtle]{savetrees}

\usepackage[T1]{fontenc}
\usepackage[scaled]{beramono}

\usepackage{pgfplots}
\usepgfplotslibrary{dateplot}
\pgfplotsset{compat=1.15}
\usepackage{anyfontsize}

\usepackage[mathscr]{eucal}
\usepackage{stmaryrd}

\usepackage{upgreek}
\newcommand{\citet}[1]
{\citeauthor{#1} ̃\shortcite{#1}}
\newcommand{\citep}{\cite}

\usepackage{multirow}
\usepackage{pgfcalendar}
\usepackage{xargs}

\usepackage[mathscr]{eucal}
\usepackage{lipsum}  

\newcommand{\overbar}[1]{\mkern 1.5mu\overline{\mkern-1.5mu#1\mkern-1.5mu}\mkern 1.5mu}

\newcommand{\Equation}[1]{Eq. (\ref{#1}) }

\newcommand{\argmax}[2]{\underset{#1}{\operatorname{argmax }}\, #2}

\newcommand{\E}[2]{\underset{#1}{\mathbb{E}}\left[ #2 \right]}

\newcommand{\w}{\mathbf{w}}
\newcommand{\x}{\mathbf{x}}
\newcommand{\y}{\mathbf{y}}

\newcommand{\FCal}{\mathscr{F}}
\newcommand{\HCal}{\mathscr{H}}

\newcommand{\SCal}{\mathscr{S}}

\newcommand{\XCal}{\mathscr{X}}

\newcommand{\Real}{\mathbb{R}}

\theoremstyle{definition}
\newtheorem{definition}{Definition}[section]

\theoremstyle{definition}
\newtheorem{theorem}{Theorem}[section]

\theoremstyle{definition}

\theoremstyle{definition}
\newtheorem{lemma}{Lemma}[section]
 
\theoremstyle{remark}

\usepackage[hyperref]{emnlp2020}
\usepackage{times}
\usepackage{latexsym}

\usepackage{microtype}

\aclfinalcopy 
\def\aclpaperid{3078} 

\title{{S}up{MMD}: A Sentence Importance Model for Extractive Summarization using Maximum Mean Discrepancy}

\author{Umanga Bista$^{\star}$\ Alexander Patrick Mathews$^{\star}$\ Aditya Krishna Menon$^{\ddag}$\ Lexing Xie$^{\star}$\\
$^{\star}$Australian National University, Canberra, ACT, Australia \\
$^{\ddag}$Google Research, New York, NY, United States  \\
$^{\star}${\tt\{umanga.bista,alex.mathews,lexing.xie\}@anu.edu.au},\\
$^{\ddag}$\tt{adityakmenon}@google.com 
}

\date{}
\begin{document}
\maketitle
\begin{abstract}
Most work on multi-document summarization has focused on \emph{generic} summarization of information present in each individual document set.
However,
the under-explored setting of \emph{update summarization},
where the goal is to identify the \emph{new} information present in each set,
is of equal practical interest (e.g., presenting readers with updates on an evolving news topic).
In this work, we present SupMMD, 
a novel technique for 
generic and update 
summarization
based on the \emph{maximum mean discrepancy} from kernel two-sample testing.
SupMMD combines both supervised learning for salience and unsupervised learning for coverage and diversity.
Further,
we adapt multiple kernel learning
to make use of similarity across multiple information sources (e.g., text features and knowledge based concepts).
We show the efficacy of SupMMD in both generic and update summarization tasks 
by meeting or exceeding the current state-of-the-art on the DUC-2004 and 
TAC-2009 datasets.
\end{abstract}

\section{Introduction}
\label{sec:intro}

\begin{figure}[t]
    \centering
    \includegraphics[width=0.48\textwidth]{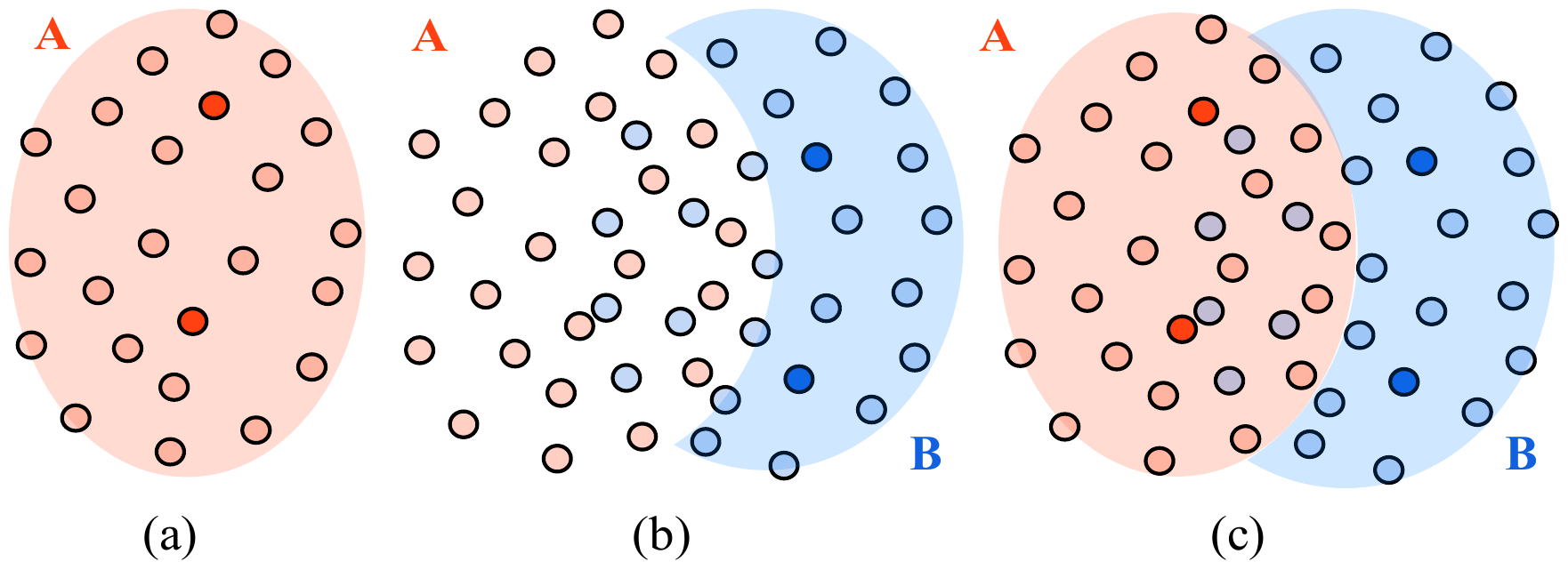}
    \caption{Different summarization tasks: (a) Generic (b) Comparative (c) Update. 
    Two sets of articles (set A and B) are denoted by red and blue circles, respectively. Summary prototypes are bold circles, and information coverage of each tasks is filled with respective colors.}
    \label{fig:summ_types_venn}
\end{figure}

Multi-document summarization is the problem of producing condensed digests of 
salient information from multiple sources, such as articles. 
Concretely, suppose we are given two sets of articles (denoted set $\mathrm{A}$ and set $\mathrm{B}$)
on a related topic (e.g., climate change, the COVID-19 pandemic), separated by publication timestamp or geographic region.
We may then identify three possible instantiations of multi-document summarization (see Figure~\ref{fig:summ_types_venn}):
\begin{enumerate}[label=(\roman*),itemsep=-3pt,topsep=0pt,leftmargin=21pt]
    \item \emph{generic summarization}, where the goal is to summarize a set ($\mathrm{A}$ or $\mathrm{B}$) individually.
    \item \emph{comparative summarization}, where the goal is to summarize a 
    set ($\mathrm{B}$) against another set ($\mathrm{A}$) while highlighting 
    the differences.
    \item \emph{update summarization}, where the goal is both generic summarization of set $\mathrm{A}$ \emph{and} comparative summarization of set $\mathrm{B}$ versus $\mathrm{A}$. 
\end{enumerate}

Most existing work on this topic has focused on the generic summarization task.
However, update summarization is of equal practical interest.
Intuitively, the comparative aspect of this setting aims to inform a user of 
new information on a topic they are already familiar with.

Multi-document extractive summarization methods can be 
unsupervised or supervised.
\emph{Unsupervised} methods typically define salience (or coverage) using a global model 
of sentence-sentence similarity.
Methods based on 
retrieval~\citep{goldstein1999summarizing}, 
centroids \citep{radev2004centroid}, 
graph centrality \citep{erkan2004lexrank},
or utility maximization~\citep{lin2010multi,Lin:2011:CSF:2002472.2002537,gillick-favre-2009-scalable} 
have been well explored.
However, sentence salience also depends on \emph{surface features} (e.g., position, length, presence of cue words);
effectively capturing these requires supervised models specific to the dataset and task. A body of work has incorporated such information through \emph{supervised learning}, 
for example based on
point processes~\citep{kulesza2012determinantal}, 
learning important words~\citep{hong2014improving}, 
graph neural networks~\citep{yasunaga-etal-2017-graph},
and
support vector regression~\citep{varma2009iiit}.
These supervised methods have either a separate model for learning and inference, leading to a disconnect between learning sentence salience and sentence selection~\citep{varma2009iiit,yasunaga-etal-2017-graph,hong2014improving}, 
or are designed specifically for generic summarization~\citep{kulesza2012determinantal}.
In this work, we propose \emph{SupMMD}, which has a single model of learning salience and inference and can be applied to generic and comparative summarization.
We make the following contributions:\\
~\textbf{(1)} We present \emph{SupMMD}, 
a novel technique for 
both generic and update summarization
that 
combines supervised learning for salience and unsupervised learning for 
coverage and diversity. 
\emph{SupMMD} has a single model for learning and inference.\\
\textbf{(2)} We adapt multiple kernel learning~\citep{cortestwostage} into our model, which allows similarity across multiple information sources 
(e.g., text features and 
knowledge based concepts) to be used.\\
\textbf{(3)} 
We show that \emph{SupMMD} meets or exceeds 
the state-of-the-art in generic and update 
summarization on the DUC-2004 and TAC-2009 datasets.

\section{Literature Review}
\label{sec:literature_review}

Multi-document summarization can be \emph{extractive}, 
where salient 
pieces of the original text such as 
sentences are selected to form the summary;
or \emph{abstractive}, where a new text is generated by paraphrasing important information. 
The former is popular as it often creates semantically and grammatically correct summaries~\citep{nallapati2017summarunner}. 
In this work, we focus on \emph{generic} and \emph{update} \emph{multi-document} summarization in the \emph{extractive} setting.

Most extractive  
summarizers have two components: 
sentence scoring and selection.
A variety of unsupervised and supervised methods have been developed for the former.
\emph{Unsupervised} sentence scorers are based on 
centroids~\citep{radev2004centroid}, graph centrality 
~\citep{erkan2004lexrank}, 
retrieval relevance~\citep{goldstein1999summarizing}, 
word statistics~\citep{nenkova2005impact}, 
topic models~\citep{Haghighi:2009:ECM:1620754.1620807}, or
concept coverage~\citep{gillick-favre-2009-scalable,Lin:2011:CSF:2002472.2002537}. 
\emph{Supervised} techniques include: 
using a graph-based neural network~\citep{yasunaga-etal-2017-graph}, 
learning sentence quality from point 
processes~\citep{kulesza2012determinantal},
combining word importances~\citep{hong2014improving}, 
combining sentence and phrase importances~\citep{cao2015ranking}, or employing a
mixture of submodular functions~\citep{lin2012learning}.

Sentence selection methods can be broadly 
categorized as 
\emph{greedy} methods~\citep{goldstein1999summarizing,radev2004centroid,erkan2004lexrank,nenkova2005impact,cao2015ranking,Haghighi:2009:ECM:1620754.1620807,hong2014improving,kulesza2012determinantal,cao2015ranking,varma2009iiit}, 
which produce approximate solutions by iteratively selecting the sentences
with the maximal score, 
or \emph{exact} integer linear programming (ILP) based methods~\citep{gillick-favre-2009-scalable,cao2015ranking}.
Some greedy methods use an objective which belongs to a special class of set 
functions called \emph{submodular 
functions}~\citep{lin2010multi,lin2012learning,Lin:2011:CSF:2002472.2002537,kulesza2012determinantal}, 
which have good approximation guarantees under greedy 
optimization~\citep{nemhauser1978analysis}.

There has been limited research into update and comparative summarization.
Notable prior work
includes maximizing concept coverage using ILP~\citep{gillick2009icsi}, learning sentence scores using a support vector regressor~\citep{varma2009iiit}, 
and temporal content filtering~\citep{zhang2009ictgrasper}. 
\citet{bista2019comparative} cast the comparative summarization problem as classification, and use MMD~\citep{gretton2012kernel}. 
In this work, we adapt their method to learn \emph{sentence importances} driven by surface features.

\section{Summarization as Classification}
\label{sec:background}
We review a perspective introduced by~\citet{bista2019comparative},
where summarization is viewed as classification,
and provide a brief introduction to Maximum Mean Discrepancy (MMD).
Both these ideas form the basis of our subsequent method. 

\subsection{Generic Summarization as Classification}
\label{sec:unsup_mmd_generic}

Let $\{V^{\mathrm{t}}\}_{\mathrm{t}=1}^T$ be $T$ topics of articles that we wish to summarize.
For a topic $t$, we wish to select \emph{summary sentences} 
$S^{\mathrm{t}}$. 
\citet{bista2019comparative}~formulated summarization as selecting prototypes
that minimize the accuracy of a powerful classifier between
sentences in the input and summary.
The intuition is that a powerful classifier should not be able to distinguish between the sentences from articles and summary sentences.
Formally, we pick
\begin{equation}
    S^{\mathrm{t}} = \argmax{S \in \SCal^t }-\mathrm{Acc}(V^{\mathrm{t}}, S),
    \label{eqn:cbc_generic}
\end{equation}
where 
$\SCal^{\mathrm{t}} \subset 2^{ V^{\mathrm{t}} } :
 \forall_{S' \in \SCal} \sum_{s \in S'} \mathrm{len}(s) \le L $
comprise subsets of $V^{\mathrm{t}}$ with upto $L$ words,
and
$\mathrm{Acc}( X, Y )$ is the accuracy of the best possible classifier that distinguishes between elements in sets $X$ and $Y$;
we shall shortly 
realize this using MMD. 

\subsection{Comparative Summarization as Competing Binary Classification}
\label{sec:unsup_mmd_comparative}

For comparative summarization between two sets $A$ and $B$,~\citet{bista2019comparative} introduced an additional term 
into~\eqref{eqn:cbc_generic}, 
giving rise to \emph{competing goals} for the classifier:
it should not be able to distinguish between the summaries and sentences from set $B$,
but \emph{should} be able to distinguish between the summaries and sentences from set $A$.
Formally, let 
$V_B^{\mathrm{t}}$ be the set of sentences in set $B$, 
$V_A^{\mathrm{t}}$ be the sentences in set to compare (set $A$).
Then, for suitable $\lambda > 0$, we seek $S^{\mathrm{t}}$, the summary sentences of set $B$,
\begin{equation}
    \resizebox{0.85\linewidth}{!}{
    $\displaystyle
    S^{\mathrm{t}} = \argmax{S \in \SCal^{\mathrm{t}} } \left[ -\mathrm{Acc}(V_B^{\mathrm{t}}, S) + \lambda \cdot \mathrm{Acc}(V_A^{\mathrm{t}}, S) \right].
    \label{eqn:cbc_comp}
    $}
\end{equation}
The hyperparameter $\lambda$ controls the relative importance of accurately 
representing 
articles in set B,
versus not representing the articles in set A.

\subsection{Maximum Mean Discrepancy (MMD)}
\label{ssec:mmd}

The 
\emph{MMD} is a kernel-based measure of the distance between two distributions.
More formally:
\begin{definition}
Let $\mathscr{H}$ be a Reproducing Kernel Hilbert Space (RKHS) 
with associated 
kernel $k$.
Let $\mathscr{F}$ be the set of functions $h: \XCal \mapsto \Real$ in
the unit ball of $\HCal$, 
where $\XCal$ is a topological space.
Then, the MMD between distributions $p, q$
is the \emph{maximal difference} in expectations of functions from $\FCal$ under $p,q$~\citep{gretton2012kernel}:
\begin{equation}
    \resizebox{0.85\linewidth}{!}{
    $\displaystyle
    \mathrm{MMD}_{\mathscr{F}}( p, q) = \sup_{h \in \mathscr{F}} \left( \E{x \sim p}{h(x)} - \E{y \sim q}{h(y)} \right). 
    \label{eqn::mmd_1}
    $}
\end{equation}
\end{definition}
{
A small MMD value indicates that $p, q$ are similar.}
Given finite samples $X \sim p^n$ and $Y \sim q^m$,
an empirical estimate of the MMD, denoted as $\mathrm{MMD}^2_{\mathscr{F}}(X, Y)$, can be computed as:
\begin{equation}
    \resizebox{0.85\linewidth}{!}{
    $\displaystyle
        \frac{1}{ n ^2}\sum_{x, x'} k(x, x') + \frac{1}{m^2}\sum_{y, y'} k(y, y') - \frac{2}{ n \cdot m}\sum_{x, y} k(x, y). \label{eqn::mmd2_biased}
    $}
\end{equation}

\subsection{MMD for Summarization}
\label{ssec:unsup_mmd_summ}

The MMD corresponds to the minimal achievable loss of a centroi-based kernel classifier~\citep{fukumizu2009kernel}.
Consequently, we use $\mathrm{MMD}^2_{\mathscr{F}}(V, S)$ to approximate the $\mathrm{Acc}(V, S)$ in~\eqref{eqn:cbc_generic} and~\eqref{eqn:cbc_comp},
{
using a suitable kernel $k$ that measures the similarity of two sentences.
Intuitively, this selects summaries $S$ which best represent the \emph{distribution} of original sentences $V$.}

Note that if we expand $\mathrm{MMD}^2_{\mathscr{F}}(V, S)$ {as per~\eqref{eqn::mmd2_biased}} and later in~\S \ref{ssec:inference}, the first term is irrelevant for optimization.
The second and third term capture the coverage and diversity of the summary sentences without any supervision.
Hence, this is an \emph{unsupervised} summarization.

\section{The SupMMD Method}
\label{sec:methods}

We start by developing a technique for incorporating sentence importance into 
MMD for the purpose of generic multi-document extractive summarization. 
We then 
extend this method to comparative summarization, 
and incorporate multiple 
different kernels to use a diverse sets of features.

\subsection{From MMD to Weighted MMD}
\label{ssec:sup_mmd}

Unsupervised MMD~\citep{bista2019comparative} selects representative sentences that cover relevant concepts 
while retaining diversity.
The notion of representativeness is based on a global model 
of sentence-sentence similarity; however, this 
notion of representativeness 
is 
not necessarily well matched to the selection of salient information.
Salience of a sentence may be determined by 
\emph{surface features} such as position in the 
article, or number of words.
For example, news 
articles
are often written such that sentences at the start of a article have the 
characteristics of a summary~\citep{kedzie-etal-2018-content}.
{Learning a notion of salience that is specific to the summarization task 
and dataset requires supervised training.
Thus, we extend the 
MMD model by incorporating supervised 
\emph{sentence importance weighting}.}

Let $v, s \in \XCal$ be independent samples drawn from the distributions 
of article sentences $p$ and summary sentences $q$ on the space of all sentences $\XCal$.
We define non-negative \emph{importance functions} 
$f^p_\uptheta$,
$f^q_\uptheta$
parameterized by 
learnable parameters $\uptheta$.
We restrict these 
functions
so 
that  
$\mathbb{E}_p{f^p_\uptheta(v)}=1$ and $\mathbb{E}_q{f^q_\uptheta(s)}=1$.
Equipped with $f_\uptheta$, we may modify MMD
such that the importance of
sentences which are good summary candidates is increased.
\begin{definition}
The \emph{weighted MMD}  
$\mathrm{MMD}_{\mathscr{F}}(p, q, \uptheta)$
between $p, q$ is
\begin{equation}
   \sup_{h \in \mathscr{F} } \left(  \E{p}{f^p_\uptheta(v) \cdot h(v)} - 
\E{q}{f^q_\uptheta(s) \cdot h(s)} \right)
   \label{eqn:sup_mmd}
\end{equation}
\end{definition}
Note that classic MMD~\eqref{eqn::mmd_1} is a special case of 
~\eqref{eqn:sup_mmd} where 
$f_{\uptheta} \equiv 1$.

{In practice, the supremum over all $h$ is impossible to compute 
directly. 
We thus derive an alternative form for Equation~\ref{eqn:sup_mmd}.}

\begin{lemma}
    \label{theorem:supmmd}
    For $\| h \|_\mathscr{H} \le 1$, 
  ~\eqref{eqn:sup_mmd} is equivalently
\begin{equation}
    \left\| \mathbb{E}_p{[f^p_\uptheta(v) \cdot \phi(v)]} - \mathbb{E}_q{[f^q_\uptheta(s)\cdot \phi(s)]} \right\|_\mathscr{H}
    \label{eqn::mmd_norm}.
\end{equation}
\end{lemma}

In the above, $\phi:\XCal \mapsto \mathscr{F}$ is a canonical feature mapping of sentences and summaries from $\XCal$ to RKHS. 
The 
derivation, which mirrors a similar derivation for 
MMD~\citep{gretton2012kernel}, is given in the 
Appendix.

\subsection{Importance Function}
\label{ssec:imp_function}

We use log-linear models as importance functions, as they are a common 
choice of sentence importance~\citep{kulesza2012determinantal}
and easy to fit when 
training data is scarce. 
Formally, the log-linear importance function is: 
$f_\uptheta(v) = \exp(\langle \uptheta, \upomega(v) \rangle) $, 
where $\upomega(v)$ is the surface features of sentence $v$. 
We can define the empirical estimates $f_\uptheta^{ n_{\mathrm{t}} }(v)$, 
$f_\uptheta^{ m_{\mathrm{t}} }(s)$ of the importance functions 
$f^p_\uptheta(v)$ and $f^q_\uptheta(s)$ as:
\begin{align}
    f_\uptheta^{n_{\mathrm{t}}}(v) =& \frac{f_\uptheta(v)}{\sum_{v' \in V^{\mathrm{t}}}f_\uptheta(v') } \cdot n_{\mathrm{t}} \nonumber \\
    f_\uptheta^{m_{\mathrm{t}}}(s) =& \frac{f_\uptheta(s)}{\sum_{s' \in S^{\mathrm{t}}}f_\uptheta(s') } \cdot m_{\mathrm{t}}
    \label{eqn:normalized_importance} 
\end{align}
{where {\small $n_{\mathrm{t}}=|V^{\mathrm{t}}|$} is
the number of sentences and {\small $m_{\mathrm{t}}=|S^{\mathrm{t}}|$} is the number 
of 
summary sentences in topic $\mathrm{t}$.
}

\subsection{Training: Generic Summarization}
\label{ssec:training}

The parameters $\uptheta$ of the log-linear importance function must be learned 
from data, 
so we define a loss function based on weighted MMD. 
Let {$\{(V^{\mathrm{t}}, S^{\mathrm{t}} )\}_{\mathrm{t}=1}^T$} be the $T$ training tuples.
Then, the loss of topic $t$ is 
the square of importance weighted 
empirical MMD between sentences and summary sentences from within the topic:
\begin{equation}
    \mathscr{L}^{\mathrm{t}}(V^{\mathrm{t}}, S^{\mathrm{t}}, \uptheta) = \mathrm{MMD}^2_{\mathscr{F}}(V^{\mathrm{t}}, S^{\mathrm{t}}, \uptheta ) \label{eqn:sup_mmd_obj} 
\end{equation}
where $\mathrm{MMD}^2_{\mathscr{F}}(V^{\mathrm{t}}, S^{\mathrm{t}}, \uptheta )$ is an 
empirical estimate of the weighted $\mathrm{MMD}^2_{\mathscr{F}}(p, q, \uptheta)$.
Applying the kernel trick to Equation~\ref{eqn:sup_mmd_obj} gives (see Appendix):
\begin{align}
    \mathscr{L}^{\mathrm{t}} =& 
    \frac{1}{n_{\mathrm{t}} ^2}\sum_{v, v'}{f_\uptheta^{n_{\mathrm{t}}}(v) \cdot  f_\uptheta^{n_{\mathrm{t}}}(v') \cdot k(v, v') } \nonumber \\
    &-\frac{2}{n_{\mathrm{t}} \cdot m_{\mathrm{t}}}\sum_{v, s}{f_\uptheta^{n_{\mathrm{t}}}(v) \cdot f_\uptheta^{m_{\mathrm{t}}}(s) \cdot k(v, s) } \nonumber \\
    &+\frac{1}{m_{\mathrm{t}}^2}\sum_{s, s'}{f_\uptheta^{ m_{\mathrm{t}} }(s) \cdot  f_\uptheta^{ m_{\mathrm{t}} }(s') \cdot k(s, s') }
    \label{eqn::sup_mmd_emp}
\end{align}
Equation \ref{eqn::sup_mmd_emp} is the loss for a single topic but during 
training we will instead minimize the average loss 
over all topics in the training set,
i.e.,
$
    \min_\uptheta \frac{1}{T}\sum_{t=1}^{T} \mathscr{L}^{\mathrm{t}}(V^{\mathrm{t}}, S^{\mathrm{t}}, \uptheta).
$
Intuitively, we learn the parameters $\uptheta$ by minimizing an 
importance weighted distance between sentences and ground truth summary sentences over all 
topics.

\subsection{Training: Comparative Summarization}
\label{ssec:training_comp}
We now extend the learning task to comparative summarization using the
competing binary classifiers idea
of~\citet{bista2019comparative}
(cf.~\S\ref{eqn:cbc_comp}). 
Specifically,
we
replace the accuracy terms in Equation~\ref{eqn:cbc_comp} with the square of 
weighted MMD. 
Given the $T$ comparative training tuples $\{(V_B^{\mathrm{t}}, 
V_A^{\mathrm{t}}, S^{\mathrm{t}})\}_{t=1}^T$, then the objective is to minimize:
\begin{equation}
    \resizebox{0.85\linewidth}{!}{
    $\displaystyle
    \min_{\uptheta_B,\uptheta_A } \frac{1}{T}\sum_t \left( \mathscr{L}^{\mathrm{t}}(V_B^{\mathrm{t}}, S^{\mathrm{t}}, \uptheta_B ) - \lambda \cdot \mathscr{L}^{\mathrm{t}}(V_A^{\mathrm{t}}, S^{\mathrm{t}}, \uptheta_A)  \right) \label{eqn:obj_train_comp}
    $}
\end{equation}
Note there are two sets of importance parameters $\uptheta_B, \uptheta_A$ one 
for each of the two document sets.

\subsection{Multiple Kernel Learning}
\label{ssec:mkl} 
We employ Multiple Kernel Learning (MKL) to make use of data from multiple sources in 
our MMD summarization framework. 
We adapt two stage kernel learning~\citep{cortestwostage}, 
where different kernels are linearly combined to maximize the alignment with 
the \textit{target kernel} of the classification problem.
Since MMD can be interpreted as classifiability~\citep{fukumizu2009kernel} MKL 
fits neatly into our MMD based summarization objective. 
Intuitively, MKL should identify a good combination of kernels 
for building a classifier that separates summary and non-summary sentences.

Let $\{k_i\}_{i=1}^p$ be $p$ kernel functions. 
For topic $t$, let $\textbf{K}_i^{\mathrm{t}}$ be the kernel matrix according to kernel function $k_i$,
and $\overbar{\textbf{K}}_i^{\mathrm{t}}=\textbf{U}_{n_{\mathrm{t}}} \textbf{K}_i^{\mathrm{t}} \textbf{U}_{n_{\mathrm{t}}}$ be the centered kernel matrix, with $\textbf{U}_{n_{\mathrm{t}}} = \textbf{I} - \textbf{11}^{\mathrm{T}}/n_{\mathrm{t}}$. 
Let $\y^{\mathrm{t}} = \{\pm 1\}^{n^{\mathrm{t}}}$ be the ground truth 
summary labels with $y_i^{\mathrm{t}}=+1$ iff $i \in S^{\mathrm{t}}$. 
The \emph{target kernel} $\y^{\mathrm{t}} (\y^{t})^{T}$ represents the ideal notion of similarity between sentences.
The non-negative kernel weights $\textbf{w}$ which lead to the optimal 
alignment 
with the target kernel
are given by%
~\citep{cortestwostage}
\begin{equation}
    \min_{\w \ge 0} \w^{\mathrm{T}} ( \textbf{M}^{t} )^{\mathrm{T}} \w - 2 \w^{\mathrm{T}} \textbf{a}^{\mathrm{t}},
\end{equation}
where $\textbf{M}^{\mathrm{t}} \in \mathbb{R}^{p\times p}$ has
$\textbf{M}_{rs}^{\mathrm{t}}=\langle \overbar{\textbf{K}}_r, \overbar{\textbf{K}}_s \rangle_{\mathrm{F}}$ and 
$\textbf{a}^{\mathrm{t}} \in \mathbb{R}^p$ has
$a_i = \langle \overbar{\textbf{K}}_i^{\mathrm{t}}, \y^{\mathrm{t}} (\y^{t})^{\mathrm{T}} \rangle_{\mathrm{F}} $.

The kernel function must be characteristic for MMD to be a valid metric~\citep{muandet2017kernel}.
Most popular kernels used for bag of words like text features (including 
TF-IDF), the
linear kernel ($k(\x,\y) = \langle \x, \y \rangle$) and 
the cosine kernel
($k(\x,\y) = \frac{\langle \x, \y \rangle}{\| \x \| \| \y \|}$), 
are not characteristic~\citep{sriperumbudur2010hilbert}. 
Fortunately, the exponential kernel, 
$k(\x, \y) = \exp(\gamma k'(\x, \y)), \ \gamma > 0$,
is characteristic for 
any kernel $k'$ 
~\citep{steinwart2001influence}.
Hence, we use the \emph{normalized exponential kernel} combined 
with the cosine kernel,
$k(\x, \y) = \exp(-\gamma)\exp(\gamma \sum_{i=1}^p w_i \cdot \cos(\x^{(i)}, \y^{(i)}))$.

\subsection{Inference}
\label{ssec:inference}

Given a learned importance function 
$f_\uptheta$,
we may 
find the best set of 
summary sentences $\bar{S}^t$ for generic summarization 
via:
\begin{equation}
    \bar{S}^{t} = \argmax{S \subset \SCal^{\mathrm{t}} 
    }-\mathscr{L}^{\mathrm{t}}(V^{\mathrm{t}}, S^{\mathrm{t}}, \uptheta) 
    \label{eqn::obj_inference_generic}
\end{equation}

Similarly, for the comparative task,
with learned importance functions,
we seek $\bar{S}^t$ as:
\begin{equation}
    \resizebox{0.825\linewidth}{!}{$
    \displaystyle
    \argmax{S \subset \SCal^{\mathrm{t}}} (- \mathscr{L}^{\mathrm{t}}(V_B^{\mathrm{t}}, S^{\mathrm{t}}, \uptheta_B) + \lambda \mathscr{L}^{\mathrm{t}}(V_A^{\mathrm{t}}, S^{\mathrm{t}}, \uptheta_A)) \label{eqn::obj_inference_comp}
    $}
\end{equation}
Both these inference problems are budgeted maximization problems, which are 
often solved by greedy algorithms~\citep{lin2010multi}. 
The generic unsupervised summarization task is submodular and monotone under 
certain conditions~\citep{kim2016examples}, so greedy algorithms have good 
theoretical guarantees~\citep{nemhauser1978analysis}. While our supervised 
variants do not have these guarantees, we find that greedy optimization 
nonetheless leads to good solutions. 

\section{Experimental Setup}
\label{sec:experiments}
We include guidance on applying \textit{SupMMD} and 
the details required to reproduce our experiments.

%
\subsection{Datasets}\label{ssec:datasets}

We use four standard multi-document summarization benchmark datasets: DUC-2003, 
DUC-2004, TAC-2008 and TAC-2009\footnote{\scriptsize{\url{https://duc.nist.gov/data.html}}};
dataset statistics 
are provided in Table~\ref{table:datasets}. Each of these datasets has multiple 
topics, where each topic in turn has multiple news articles and four human 
written summaries. 
In one setting we use DUC-2003 as the training set and DUC-2004 
as test set, and in another setting we use TAC-2008 as the training set and TAC-2009 as the 
test set -- both settings are common in the literature. The DUC datasets can be used for generic  
summarization while TAC, 
being 
an update summarization task, can be used for both 
generic 
(set A) and 
comparative 
summarization
(set B).
\begin{table*}[t]
    \centering
    {\small
    \begin{tabular}{lrr|rrr|rrr}
    \toprule
              &        &      & \multicolumn{3}{c|}{Oracle (ours)}   & \multicolumn{3}{c}{Oracle \footnotesize{\citep{liu-lapata-2019-text} }} \\
    \bf Dataset  &\bf \# topics &\bf \# sents &\bf avg summ sents 
    &\bf R1 &\bf R2 &\bf avg summ sents &\bf R1 &\bf R2  \\
    \midrule
    DUC2003   & 30     & 6989  & 3.73        & 43.1 & 17.0 & 3.40       & 42.2 & 16.2  \\
    DUC2004   & 50     & 12148 & 4.02        & 42.0 & 14.9 & 3.46       & 40.6 & 14.2  \\
    TAC2008-A & 48     & 9914  & 3.90        & 45.5 & 19.4 & 3.42       & 44.0 & 18.6  \\
    TAC2008-B & 48     & 9147  & 3.83        & 44.9 & 19.5 & 3.50       & 43.6 & 18.7  \\
    TAC2009-A & 44     & 9509  & 4.07        & 46.9 & 20.5 & 3.32       & 44.5 & 19.1  \\
    TAC2009-B & 44     & 8543  & 3.61        & 44.8 & 19.2 & 3.27       & 43.1 & 18.1  \\
    \bottomrule    
    \end{tabular}
    }

    \caption{Dataset statistics and oracle performance. 
    We report the number of topics in each dataset, 
    along with the number of sentences after preprocessing. 
    We show the ROUGE scores of our oracle method and the one by \citet{liu-lapata-2019-text} with average number of sentence in summary from each method.
    }
    \label{table:datasets}
    \end{table*}

\subsection{Data Preprocessing and Preparation}
\label{ssec:experiments}

The DUC and TAC datasets are provided as collections of XML documents, so 
it is necessary to 
extract relevant text
and then perform sentence and word tokenization.
For DUC we clean the text using various regular expressions the details of 
which are provided in our code release. 
We train {\small\texttt{PunktSentenceTokenizer}} to detect 
sentence boundaries, and use the standard NLTK~\citep{nltk} word tokenizer. 
For the TAC dataset, we 
use the preprocessing pipeline employed by~\citet{gillick2009icsi} 
\footnote{\scriptsize{\url{https://github.com/benob/icsisumm}}}.
This enables a cleaner comparison with the 
state-of-the-art ICSI~\citep{gillick2009icsi} method on the TAC dataset.
For all datasets, we 
keep the sentences between 8 and 55
words per~\citet{yasunaga-etal-2017-graph}.

\subsection{Feature Representations}
\label{ssec:features_rep} 
Our method requires two different sets of sentence features:
\emph{text features}, which are used to compute the sentence-sentence similarity as part 
of the kernel;
and \emph{surface features}, which are used in 
learning the sentence importance model.

\subsubsection{Text Features}
\label{ssec:text_feats}

Each sentence has three different feature representations: unigrams, bigrams and entities. The unigrams are stemmed words, with stop 
words 
from the NLTK english list
removed. 
The bigrams are a combination 
of stemmed unigrams and bigrams.
The entities are 
DBPedia concepts extracted using DBPedia Spotlight~\citep{mendes2011dbpedia}.

We use a Term Frequency Inverse Sentence Frequency (TF-ISF)~\citep{neto2000document}
representation for all text features.
TF-ISF has been used extensively in 
multi-document summarization~\citep{dias2007topic,alguliev2011mcmr,wan2007towards}.

\subsubsection{Surface Features}
\label{ssec:surf_feats}
  
We use $10$
surface features for the DUC 
dataset, and $12$ for the TAC dataset:    
 
\noindent\textbf{position}: There are five position features. 
Four indicators 
denote the $1^{\mathrm{st}}$, $2^{\mathrm{nd}}$, $3^{\mathrm{rd}}$ or a later position of the sentence in the article.
The final feature gives the position relative to the length 
of the article.

\noindent\textbf{counts}: There are two count features: the number of words and 
number of nouns. We use the {\small{\tt spaCy}}~\footnote{\scriptsize{\url{https://spacy.io}}} 
part 
of speech tagging to find nouns. 

\noindent\textbf{tfisf}: This is the sum of the TS-ISF scores for unigrams composing the 
sentence. For sentence $s$, this is $\sum_{w \in s}{\mathrm{isf}(w) 
\cdot \mathrm{tf}(w, s)}$, where 
$\mathrm{isf}(w)$ 
is the inverse sentence frequency of unigram $w$, and $\mathrm{tf}(w, s)$ is the 
term frequency of $w$ in $s$.

\noindent\textbf{btfisf}: The boosted sum of TS-ISF scores for unigrams composing 
the sentence. 
Specifically, we compute $\sum_{w 
\in s}{\mathrm{isf}(w) \cdot b(w) \cdot \mathrm{tf}(w, s)}$, where
we boost the score of unigrams $w$ 
that appear in the first sentence of the article as $b(w)$.  
In the generic summarization 
$b(w)=2$,
for comparative summarization 
$b(w)=3$, as used by \citet{gillick2009icsi}. 
Unigrams that do not appear in the first sentence of the article have
$b(w) = 1$.
    
\noindent\textbf{lexrank} The LexRank score~\citep{erkan2004lexrank} computed 
on the bigrams' cosine similarity.

For the TAC datasets, we additionally use:

\noindent\textbf{par\_start}: An indicator whether the sentence begins a paragraph. 
This is provided by the preprocessing pipeline from ICSI~\citep{gillick2009icsi}.

\noindent\textbf{qsim}: The fraction of topic description unigrams present in each sentence;
these topic descriptions are only available for TAC.

\subsection{Oracle Extraction}
\label{ssec:oracle}

Both DUC and TAC provide four human written 
summaries for each topic. 
Since our goal is extractive summarization with supervised training,
we need to 
know which 
sentences in the articles 
could be used to construct the summaries 
in the training set. 
The article sentences that best match the abstractive summaries are called 
the \emph{oracles} ($S^\mathrm{t}$).
\begin{algorithm}
    \caption{Oracle extraction}
    \label{algo:oracle}
    \begin{algorithmic}[1]
      \Function{ExtractOracle}{$\alpha, V^\mathrm{t}, H^\mathrm{t}, r, L$}
      \State $S^\mathrm{t} \gets \emptyset$
      \While{$\sum_{s \in S^\mathrm{t}}\mathrm{len}(s) \le L$}
      \State $ s^* \gets \argmax{s \in V^\mathrm{t} \setminus S^\mathrm{t} } {\frac{\alpha(S^\mathrm{t} \cup \{s\}, H^\mathrm{t}) - \alpha(S^\mathrm{t}, H^\mathrm{t}) }}{\mathrm{len}(s)^r} $
      \State $S^\mathrm{t} \gets S^\mathrm{t} \cup \{s^*\}$
      \EndWhile
      \Return{$S^t$}
      \EndFunction
    \end{algorithmic}
\end{algorithm}

Our extraction algorithm 
(Algorithm \ref{algo:oracle}), is 
inspired by 
\citet{liu-lapata-2019-text}.
We greedily select sentences ($s$) which provide the maximum 
gain in extraction score $\alpha(S^\mathrm{t}, H^\mathrm{t})$ against the human summaries ($H^{\mathrm{t}}$) until a 
word budget ($L$) is 
reached. 
We only include sentences between 8 to 55 words as suggested by 
\citet{yasunaga-etal-2017-graph}, 
and set a budget of 104 words to ensure our 
oracle summaries are within $100\pm 4$ words, 
consistent with the evaluation (~\S\ref{ssec:evaluation}). 

In contrast to~\citet{liu-lapata-2019-text} which uses only 
ROUGE-2 recall score~\citep{lin-2004-rouge}, our method balances both ROUGE-1 and 
ROUGE-2 recall scores using the 
harmonic mean
and explicitly accounts for sentence 
length. 
Grid search on the 
validation sets shows that the optimal value for $r$ is $0.4$ across different 
datasets and summarization tasks.
As reported in Table~\ref{table:datasets}, 
on average our method produces oracles consisting of more sentences and with
higher 
ROUGE-1 and ROUGE-2 
scores compared to oracles from \citet{liu-lapata-2019-text}. 
This is consistent across all datasets.

\subsection{Implementation Details}
\label{ssec:implmentation}
Supervised variants use an $\ell_2$ regularized log-linear model 
of importance (\S\ref{ssec:imp_function}) trained using the oracles 
(\S\ref{ssec:oracle}) as ground truth. 
{We selected the number of training epochs using 5-fold cross
validation.}
We then tune the other hyperparameters {on the training set}. 
The hyperparameters of the generic summarization task are: $\gamma$, a parameter of 
the kernel;
$\beta$, the $\ell_2$ regularization weight for the log-linear 
importance function; and $r$, which defines the length dependent scaling 
factor in greedy selection~\citep{lin2010multi}. 
The comparative objective \eqref{eqn:obj_train_comp} has an additional hyperparameter $\lambda$, which controls 
the comparativeness. 
More implementation details are provided in the Appendix.
We will make implementation publicly available\footnote{\url{github.com/computationalmedia/supmmd}}.

\subsection{Evaluation Settings}\label{ssec:evaluation}
To evaluate our methods we use 
the ROUGE~\citep{lin-2004-rouge} metric,
the \emph{de facto} choice for evaluating
both generic summarization~\citep{hong2014improving,cho-etal-2019-improving,yasunaga-etal-2017-graph,kulesza2012determinantal},
and update summarization~\citep{varma2009iiit,gillick-favre-2009-scalable,zhang2009ictgrasper,li2009tac}.
ROUGE metrics have been shown to correlate with human judgments~\citep{lin-2004-rouge} in generic summarization task. 
Our recent work~\citep{bista2019comparative} shows that human judgments are consistent with the automatic metrics for evaluating comparative summaries.

Both DUC and TAC evaluations use the first 100 words of the generated summary. 
Our DUC-2004 evaluation setup mirrors~\citet{hong2014repository}.
This allows us to compare  
performance with the state-of-the-art methods they reported and other works also evaluated using 
this setup\footnote{\scriptsize ROUGE 1.5.5 with args -n 4 -m -a -l 100 -x -c 95 -r 1000 -f A -p 0.5 -t 0}. 
As is standard for the DUC-2004 datasets, we report 
ROUGE-1 and ROUGE-2 recall scores.

For TAC-2009 datasets 
(both set A and B)
, we adopt the evaluation settings from the TAC-2009 competition\footnote{\scriptsize{\url{tac.nist.gov/2009/Summarization}}} so we can 
compare against the three best performing 
systems in the competition\footnote{\scriptsize args -n 4 -w 1.2 -m -2 4 -u -c 95 -r 1000 -f A -p 0.5 -t 0 -a -l 100}.
As is standard for the TAC-2009 
dataset,
we report ROUGE-2 and ROUGE-SU4 recall scores.

\subsection{Baselines}
\label{ssec:baselines}
\textbf{DUC-2004}: 
We select the top performing methods from a recent benchmark paper~\citep{hong2014repository} to serve as baselines and report ROUGE scores from the benchmark paper.
They are:

\noindent\textit{ICSI}: an integer linear programming method that 
maximizes coverage~\citep{gillick2009icsi}, 

\noindent\textit{DPP}: a determinantal point process method that learns sentence quality 
and maximizes diversity~\citep{kulesza2012determinantal}, 

\noindent\textit{Submodular}: a method based on a learned mixture of submodular 
functions~\citep{lin2012learning}, 

\noindent\textit{OCCAMS\_V}: a method base on topic modeling~\citep{conroy-etal-2013-multilingual},

\noindent\textit{Regsum}: a method that focuses on learning word importance~\citep{hong2014improving},

\noindent\textit{Lexrank}: a popular graph based sentence scoring method~\citep{erkan2004lexrank}.

We also include recent deep learning methods evaluated using the same setup as~\citet{hong2014repository} and report ROUGE scores from the individual papers: 

\noindent\textit{DPP\-Sim}: an extension to the DPP model which learns the sentence-sentence
similarity using a capsule network~\citep{cho-etal-2019-improving},

\noindent\textit{HiMAP}: a recurrent neural model that employs a modified 
pointer-generator component~\citep{fabbri2019multi}, and

\noindent\textit{GRU+GCN}: a model that uses a graph convolution 
network combined with a recurrent neural network to learn sentence
saliency~\citep{yasunaga-etal-2017-graph}.

\textbf{TAC-2009}:
As baselines for the TAC-2009 dataset we use the top three systems in the 
TAC-2009 competition for each task, resulting in four systems altogether.
To the best of our knowledge these systems are the current state-of-the-art. 
We report the ROUGE scores from the competition.
The systems are:

\noindent\textit{ICSI}: with two variants: \textit{Sys.34} uses integer linear programming to maximize 
coverage of concepts~\citep{gillick2009icsi}, and \textit{Sys.40}, which additionally uses 
sentence compression to generate new candidate sentences, 

\noindent\textit{IIT}: uses a support vector regressor to predict sentence ROUGE scores~\citep{varma2009iiit}, 

\noindent\textit{ICTCAS}: a temporal content filtering method~\citep{zhang2009ictgrasper}, 
and 

\noindent\textit{ICL}: a manifold ranking based method~\citep{li2009tac}.

\section{Experimental Results}
\label{sec:results}

We compare our methods with the baselines 
on the 
DUC-2004, TAC-2009-A and TAC-2009-B datasets. 
We present several variants of our method to 
analyze the effects of different components and modeling choices.
We report the performance of unsupervised MMD (\textit{UnsupMMD}) which 
does not explicitly consider sentence importance.
For our supervised method \textit{SupMMD}, 
we report the performance with a bigram kernel (\textit{SupMMD}) and combined 
kernels (\textit{SupMMD + MKL}). We also evaluated the impact of our oracle 
extraction method by replacing it with the extraction method suggested by 
\citet{liu-lapata-2019-text} in \textit{SupMMD + alt oracles}. Meanwhile, 
\textit{SupMMD + MKL + compress} presents the result of applying sentence 
compression~\citep{gillick2009icsi} to our model.
\subsection{Generic Summarization}
\label{ssec:res_generic_summary}
\begin{table}[ht]
    \centering
    {
    \small
    \begin{tabular}{l|rr}
    \toprule
    \textbf{DUC-2004}                         & \textbf{R1}                      & \textbf{R2}            \\
    \midrule
    \textit{ICSI} \footnotesize{\citep{gillick2009icsi}}& 38.41          & 9.78 \\
    \textit{DPP} \footnotesize{\citep{kulesza2012determinantal}}             & \textbf{39.79} & 9.62 \\
    \textit{Submodular} \footnotesize{\citep{lin2012learning}}& 39.18 & 9.35 \\
    \textit{OCCAMS\_V} \footnotesize{\citep{conroy-etal-2013-multilingual}}& 38.50 & 9.76 \\
    \textit{Regsum} \footnotesize{\citep{hong2014improving}}& 38.57 & 9.75 \\
    \textit{Lexrank} \footnotesize{\citet{erkan2004lexrank}}& 35.95 & 7.47 \\
    \textit{DPP-Sim} \footnotesize{\citep{cho-etal-2019-improving}}& 39.35 & 10.14 \\
    \textit{HiMAP} \footnotesize{\citep{fabbri2019multi}}& 35.78 & 8.90 \\
    \textit{GRU+GCN} \footnotesize{\citep{yasunaga-etal-2017-graph}}& 38.23 & 9.48 \\
    \midrule
    UnsupMMD       & 35.73      & 7.76     \\
    SupMMD (alt oracle)   & 39.02     & 10.22   \\
    SupMMD      & 39.36      & 10.31         \\
    SupMMD + MKL + compress     & \it 39.63  &\it  10.50         \\
    SupMMD + MKL          & 39.27      & \textbf{10.54} \\
    \bottomrule           
    \end{tabular}
    }
    \caption{Results on DUC-2004 generic multi-document summarization task.}
    \label{table:duc}
\end{table}
The performance of our methods on the DUC-2004 generic 
summarization task are shown in Table~\ref{table:duc}. 
On the DUC-2004 dataset all \textit{SupMMD} variants exceed the 
state-of-the-art, 
when evaluated with 
ROUGE-2, and perform similarly to the best existing methods when evaluated with 
ROUGE-1. 
Our best system \textit{SupMMD + MKL} outperforms the previous best system 
(\textit{ICSI}) on ROUGE-2 score by $\mathbf{+3.9}\%$. While the \textit{DPP} 
baseline achieves the highest ROUGE-1 score on DUC-2004, it has a relatively 
low ROUGE-2 score which suggests it is optimized for unigram performance 
at the cost of bigram performance. 
\textit{SupMMD + MKL} strikes a better 
balance, scoring the best in ROUGE-2 and second best in ROUGE-1.
On the 
TAC-2009 generic 
summarization task in Table \ref{table:tac-A} our \textit{SupMMD + MKL} model 
outperforms the state-of-the-art ICSI model 
on both ROUGE-2 and ROUGE-SU4.
Specifically, \textit{SupMMD + MKL} scores 12.33 in 
ROUGE-2 while the best ICSI variant scores 12.16 in ROUGE-2.

{\textbf{Supervised Modeling}}: Models using supervised training to identify important sentences 
substantially outperform the unsupervised method \textit{UnsupMMD}. 
In fact, \textit{UnsupMMD} is the lowest scoring method across all metrics and datasets. 
This strongly indicates that a degree of supervision is essential to perform 
well in this task, and that the importance function
is a suitable way to adapt the \textit{UnsupMMD} model 
to supervised training.
{Moreover, we observe a strong correlation between the the relative position of a sentence and the score given by \emph{SupMMD}. This observation is consistent with previous works~\citep{kedzie-etal-2018-content}, and demonstrates that \emph{SupMMD} has learned to use the surface features to capture salience. Further details of feature correlations are provided in the Appendix.
}

\textbf{Oracle extraction}: Our oracle extraction technique for transforming abstractive training data to extractive training data helps 
\textit{SupMMD} methods achieve higher ROUGE performance. An alternative 
technique developed by~\citet{liu-lapata-2019-text} and implemented in 
\textit{SupMMD (alt oracle)} gives lower performance than our technique.
For 
example, on DUC-2004 \textit{SupMMD (alt oracle)} has a ROUGE-1 of 39.02 and 
ROUGE-2 of 10.22, while \textit{SupMMD} has a ROUGE-1 of 39.36 and a ROUGE-2 of 
10.31. Thus, the advantages of our proposed oracle extraction method are 
substantial and consistent across multiple datasets and evaluation metrics.

\textbf{Multiple Kernel Learning}: We observe that combining multiple kernels helps the performance of
\textit{SupMMD} models 
on the generic summarization task.
\textit{SupMMD + MKL} which 
combines both bigram and entity kernels has a ROUGE-2 of 10.54 on DUC-2004, while \textit{SupMMD} only uses the bigrams kernel and scores 10.31 in 
ROUGE-2. Multiple kernels show even clearer gains in the TAC-2009-A
dataset.
\begin{table}[!t]
    \centering
    {
    \small
    \begin{tabular}{l|rr}
    \toprule
    \textbf{TAC-2009-A}           & \textbf{R2}   & \textbf{RSU4}  \\
    \midrule
    \textit{ICSI}\footnotesize{(Sys.34)} \footnotesize{\citep{gillick2009icsi}}               & 12.10          & \it 15.09 \\
    \textit{ICSI}\footnotesize{(Sys.40)} \footnotesize{\citep{gillick2009icsi}}               & \it 12.16          & {15.03} \\
    \textit{IIIT}\footnotesize{(Sys.35)} \footnotesize{\citep{varma2009iiit}}               & 10.89         & 14.49          \\
    \textit{ICTCAS}\footnotesize{(Sys.45)} \footnotesize{~\citep{zhang2009ictgrasper}}               & 10.64         & 13.99          \\

    \midrule
    UnsupMMD      & 8.35           & 11.75          \\
    SupMMD (alt oracle)  & 11.13          & 14.22    \\
    SupMMD     & 11.76          & 14.67          \\
    SupMMD + MKL + compress    & 12.02          & 15.02           \\
    SupMMD + MKL         & \textbf{12.33} & \textbf{15.19} \\


    \bottomrule          
    \end{tabular}
    }
    \caption{Results on TAC-2009 generic multi-document summarization task (TAC-2009 set A).}
    \label{table:tac-A}
\end{table}

\textbf{Sentence compression}
incorporated into the post-processing steps of \textit{SupMMD + MKL + compress} does not
clearly improve the results over \textit{SupMMD + MKL}. On TAC-2009-A, compression clearly reduces performance, 
and on DUC-2004 \textit{SupMMD 
+ MKL + compress} has a higher ROUGE-1 score but a lower ROUGE-2 score than 
\textit{SupMMD + MKL}. 
Incorporating compression into the summarization pipeline is an appealing
direction for future work.

\subsection{Comparative Summarization}\label{ssec:res_update_summary}

The results for the comparative summarization task on the TAC-2009-B dataset are 
shown in Table~\ref{table:tac-B}. Our supervised MMD variants \textit{SupMMD} 
and \textit{SupMMD + MKL} both outperform the state-of-the-art baseline ICSI in 
ROUGE-SU4 but fall short in ROUGE-2. It would be hard to claim that either 
method is superior in this instance; however, it does show that \textit{SupMMD} 
-- which uses a substantially different approach to that of \textit{ICSI} -- 
provides an alternative state-of-the-art. Thus \textit{SupMMD} further maps out 
the set of techniques that are useful for comparative summarization.
As per the generic summarization task, both our supervised training method 
and oracle extraction method are essential for achieving good 
performance in ROUGE-2 and ROUGE-SU4. We also identify sentence position and btfisf as important features for sentence salience (see the Appendix).

\textbf{Multiple kernels} as in 
\textit{SupMMD + MKL} has relatively little effect, reducing the ROUGE-2 score 
to 10.24 from the slightly higher 10.28 achieved by \textit{SupMMD}. A similar 
small decrease is seen for ROUGE-SU4. 
Manual inspection shows that the summaries from \textit{SupMMD} and \textit{SupMMD + MKL} methods are largely identical with differences primarily on topic \textsc{D0908}, which covers political movements in Nepal. 
The key entities in this topic are not resolved accurately by DBPedia Spotlight, contributing additional noise and affecting the MKL approach.

\textbf{Model variants}: We have tested an additional variant of our model for comparative 
summarization, \textit{SupMMD$^2$}, 
which defines two different importance functions: one for each of the two 
document 
sets - \textsc{A} and \textsc{B} (See \S\ref{ssec:training_comp} for details).  
In contrast, \textit{SupMMD} has a single importance function shared between document sets, i.e., in Equation \eqref{eqn:obj_train_comp}, $\uptheta_A = \uptheta_B$.
\textit{SupMMD$^2$} performed substantially worse than \textit{SupMMD} in both 
metrics, for example, \textit{SupMMD} has a ROUGE-2 of 10.28 while 
\textit{SupMMD$^2$} has a ROUGE-2 of 9.94. 
We conjecture that a single 
importance function performs better when training data is
relatively scarce because it reduces the number of parameters and simplifies 
the learning problem. Techniques for tying together the parameters for both 
importance functions, such as with a hierarchical Bayesian model, are left as 
future work.

\begin{table}[!t]
    \centering
    {
    \small
    \begin{tabular}{l|rr}
    \toprule
    \textbf{TAC-2009-B}              & \textbf{R2}    & \textbf{RSU4}  \\
    \midrule
    \textit{ICSI} \footnotesize{(Sys.34)} \footnotesize{\citep{gillick2009icsi}} & \textbf{10.39} & 13.85 \\
    \textit{ICSI} \footnotesize{(Sys.40)} \footnotesize{\citep{gillick2009icsi}} & {\it 10.37} & 13.97 \\
    \textit{IIIT} \footnotesize{(Sys.35)} \footnotesize{\citep{varma2009iiit}}            & 10.10           & 13.84          \\
    \textit{ICL} \footnotesize{(Sys.24)} \footnotesize{~\citep{li2009tac}}            & 9.62           & 13.52  \\
    \midrule
    UnsupMMD              & 7.20         & 11.29          \\
    SupMMD (alt oracle)          & 10.06        & 13.86     \\
    SupMMD$^{2}$       & 9.94         & 13.76         \\
    SupMMD       & \it 10.28        & \textbf{14.09}  \\
    SupMMD  + MKL + compress            & 10.25        & 13.91         \\
    SupMMD + MKL           & 10.24        & \it 14.05         \\
    \bottomrule
    \end{tabular}
    }

    \caption{Results on TAC-2009 comparative multi-document summarization task (TAC-2009 set B).}
    \label{table:tac-B}
\end{table}

\section{Conclusion}
\label{sec:conclusion}

In this work, we present SupMMD, 
a novel technique for update summarization
based on the \emph{maximum mean discrepancy}.
SupMMD combines supervised learning for salience, and unsupervised learning for coverage and diversity.
Further,
we adapt multiple kernel learning to exploit  multiple sources of similarity (e.g., text features and knowledge based concepts).
We show the efficacy of SupMMD in both generic and update summarization tasks on two standard datasets, when compared to the existing approaches.
We also show that the importance model we introduce on top of our existing unsupervised MMD~\citep{bista2019comparative} improves the summarization performance substantially on both generic and comparative summarization tasks.

For future work, we leave the task of incorporating embeddings features such as BERT~\citep{devlin2018bert}, and evaluating with large generic multi-document summarization dataset MultiNews~\citep{fabbri2019multi}.

\section*{Acknowledgments}
This work is supported in part by Data to Decisions CRC and ARC Discovery Project DP180101985. 
This research is also supported by use of the NeCTAR Research Cloud, 
a collaborative Australian research platform supported by the National Collaborative Research Infrastructure Strategy.
We thank Minjeong Shin for helpful feedback and suggestions.

\bibliographystyle{acl_natbib}
\bibliography{paper_3078}

\appendix
\section{Background Theory on Kernels and MMD}
\label{sec:appendix_bg}
In this section, we provide a brief overview of kernels and Maximum mean Discrepancy (MMD).
For a detailed overview, we refer readers to to~\citet{muandet2017kernel} and~\citet{gretton2012kernel} from which this brief overview is taken.
\subsection{Positive Definite Kernels and Kernel Trick}
\label{ssec:pd_kernel}
\begin{definition}
    A function $k : \XCal \times \XCal \mapsto \Real$ is called positive definite kernel if it is symmetric, i.e. $\forall_{x,y \in \XCal}\ k(x, y) = k(y, x)$ and gram matrix is positive definite, i.e.
    $ \forall_{n \in \mathbb{N}}$ $\forall_{c_1, c_2, ..c_n \in \mathbb{R}}$ $\sum_{i, j = 1}^{n} c_i c_j k(x_i, x_j) \ge 0$.
\end{definition}
\begin{theorem}\label{kernel_trick}
    If a kernel is positive definite, there exists a feature map $\phi: \XCal \mapsto \mathscr{H}$ such that $\forall_{x,y\in \XCal}\ k(x, y) = \langle \phi(x),\ \phi(y) \rangle_\mathscr{H}$.
\end{theorem}
This is known as the kernel trick in machine learning.
The feature space $\mathscr{H}$ is called a Reproducing Kernel Hilbert Space (RKHS),
and the kernel $k$ is also known as reproducing kernel.

\subsection{Reproducing Kernel Hilbert Space}
\label{ssec:rkhs} 
\begin{definition}
\label{defn:kernel_trick}
An RKHS is a Hilbert space of functions where all function evaluations are bounded, i.e. $\forall_{x \in \XCal}\ \forall_{h \in \mathscr{H}}\ \exists_{c > 0}\ \left| h(x) \right| \le c \|h\|_\mathscr{H}$.
\end{definition}
In an RKHS, function evaluation $h(x) = \langle h,\ \phi(x) \rangle_\mathscr{H}$,
where $\phi : \XCal \mapsto \mathscr{H}$ are canonical feature map associated with RKHS $\mathscr{H}$,
and $\phi(x) = k(., x)$.
A RKHS is fully characterized by its reproducing kernel $k$,
or a positive definite kernel $k$ uniquely determines a RKHS and vice versa 
.
Hence, $ \mathbb{E}_{p}{[h(x)]} = \left\langle h,\ \mathbb{E}_{p}{[\phi(x)]} \right\rangle_\mathscr{H} $,
which is known as the
Riesz representer theorem~\citep{Conway:1990}.

\subsection{More on MMD}
\label{ssec:mmd_elaborate}
Recall that $\mathscr{F}$ is a class of RKHS functions within the unit ball, i.e. $h \in \mathscr{H}, \| h \|_{\mathscr{H}} \le 1$. Suppose $\mathscr{H}$ admits a \emph{feature map} $\phi \colon \XCal \mapsto \mathscr{H}$.
Then, per~\citet{gretton2012kernel}, we may solve the supremum in Equation \ref{eqn::mmd_1} as
\begin{equation}
    \mathrm{MMD}_{\mathscr{F}}( p, q) = \left\| \mathbb{E}_{p}{ \phi(x) } - \mathbb{E}_{q}{\phi(y)} \right\|_{\mathscr{H}}. \label{eqn::mmd_2}
\end{equation}
Hence, MMD is computed as the distance between the \emph{mean feature embeddings} under each distribution, for a suitable kernel-based feature space~\citep{gretton2012kernel}. 

\Equation{eqn::mmd_2} involves explicitly evaluating the arbitrarily high-dimensional features. 
Instead, the \emph{kernel trick}
allows efficient computation of $\mathrm{MMD}^2_{\mathscr{F}}( p, q)$ by evaluating just pairwise kernels.
Supposing $\HCal$ has induced kernel $k$, we have

{\small{
\begin{align}
    \mathrm{MMD}^2_{\mathscr{F}}(p, q) =& \E{x,x' \sim p}{k(x,x')} + \E{y,y' \sim q}{k(y,y')}\nonumber \\
    -& 2 \E{x \sim p, y\sim q}{k(x,y)}.
    \label{eqn::mmd2}
\end{align}
}}

\subsection{Characteristic Kernel}
\label{ssec:char_kernels}
For a distribution $p$,
and kernel with feature map $\phi \colon \XCal \mapsto \HCal$,
the \emph{kernel mean map} is
$$ \upmu_p = \mathbb{E}_{x \sim p}\left[ \phi( x ) \right]. $$
A kernel $k$ is characteristic if the map $\upmu: p \mapsto \upmu_p $ is injective. 
A characteristic kernel ensures MMD is $0$ 
if and only if $p = q$, i.e., 
no information is lost in mapping the distribution into the RKHS~\citep{muandet2017kernel}.

Examples of characteristic kernels for $\mathbb{R}^d$ include
the
Gaussian kernel 
( $k(\x, \y) = \exp(-\gamma \| \x -\y \|_2^2), \ \gamma > 0$ ), 
and 
Laplace kernel 
( $k(\x, \y) = \exp(-\gamma \| \x -\y \|_1), \ \gamma > 0$ )
.
MMD with the Gaussian kernel is equivalent to comparing all moments between two distributions \citep{li2015generative}.

\section{Proof of Lemma~\ref{theorem:supmmd}}
\label{proof::lemma1}

The
weighted MMD $\mathrm{MMD}_{\mathscr{F}}(p, q, \uptheta)$ (\S\ref{ssec:sup_mmd}), where $\mathscr{F}$  contains functions $h:\XCal \mapsto \Real$ within unit ball RKHS $\mathscr{H}$ ($\|h\|_{\mathscr{H}} \le 1$) is defined as:
\begin{equation}
    \sup_{h \in \mathscr{F} } \left(  \E{v \sim p}{ f_\uptheta^p(v) \cdot h(v) } - \E{s \sim q}{f_\uptheta^q(s) \cdot h(s)} \right) \nonumber 
\end{equation}
Recall $f_\uptheta$ is a non-negative importance weighting function. Then, according to \citet{patil1978weighted}, the weighted probability density $\bar{p}_{\uptheta}$ of $p$ is:
\begin{equation}
\bar{p}_{\uptheta}(v) = \frac{f_\uptheta^p(v)\cdot p(v)}{\mathbb{E}_p {[f_\uptheta^p(v)]} } \nonumber
\end{equation}
and similarly $\bar{q}_{\uptheta}$ for $q$. Since we restrict $\mathbb{E}_p{[f_\uptheta^p(v)]} = 1$, and $\mathbb{E}_q{[f_\uptheta^q(s)]} = 1$, we have $\bar{p}_{\uptheta}(v) = f_\uptheta^p(v)\cdot p(v)$ and $\bar{q}_{\uptheta}(s) = f_\uptheta^q(s)\cdot q(s)$. 
Thus, the weighted MMD is
\begin{align}
    &\sup_{h \in \mathscr{F} } \left(  \E{v \sim \bar{p}_{\uptheta}}{h(v)} - \E{s \sim \bar{q}_{\uptheta}}{h(s)} \right) \nonumber \\
    &= \sup_{ || h ||_\mathscr{H} \leq 1 }  \left( \E{v \sim \bar{p}_{\uptheta}}{h(v)} - \E{s \sim \bar{q}_{\uptheta}}{h(s)} \right) \nonumber
\end{align}
Since in an RKHS, $ \mathbb{E}_{p}{[h(x)]} = \left\langle h,\ \mathbb{E}_{p}{[\phi(x)]} \right\rangle_\mathscr{H} $, this simplifies to:
\begin{align}
    &\sup_{ ||h ||_\mathscr{H} \leq 1 } \left\langle h, \E{v \sim \bar{p}_{\uptheta} }{\phi(v)} - \E{s \sim \bar{q}_{\uptheta}}{\phi(s)} \right\rangle_{\mathscr{H}} \nonumber \\
    &= \left\| \E{v \sim \bar{p}_{\uptheta}}{\phi(v)} - \E{s \sim \bar{q}_{\uptheta}}{\phi(s)} \right\|_\mathscr{H} \nonumber\\
    &= \left\| \E{v \sim p}{f^p_{\uptheta}(v) \cdot \phi(v)} - \E{s \sim q}{f^q_{\uptheta}(s) \cdot \phi(s)} \right\|_\mathscr{H}, \nonumber
\end{align}
where the penultimate step follows from the dual norm theorem\footnote{\label{dual_norm}https://en.wikipedia.org/wiki/Dual\_norm}.
The proof is similar to MMD in \citep{gretton2012kernel}.

\section{Empirical estimate of $\mathrm{MMD}^2_{\mathscr{F}}(p, q, \uptheta)$}
\label{ssec:emp_loss_appendix}
First, $\mathrm{MMD}^2_{\mathscr{F}}(p, q, \uptheta)$ can be expanded as:
\begin{align}
    & \left\| \E{v \sim p}{f^p_{\uptheta}(v) \cdot \phi(v)} - \E{s \sim q}{f^q_{\uptheta}(s) \cdot \phi(s)} \right\|^2_\mathscr{H} \nonumber \\
    =& \E{v, v' \sim p}{f^p_{\uptheta}(v) \cdot f^p_{\uptheta}(v') \cdot \langle \phi(v), \phi(v') \rangle_{\mathscr{H}} } \nonumber \\
    & - 2\cdot \E{v \sim p, s \sim q}{f^p_{\uptheta}(v) \cdot f^q_{\uptheta}(s) \cdot \langle \phi(v), \phi(s) \rangle_{\mathscr{H}} } \nonumber \\
    & + \E{s, s' \sim q}{f^q_{\uptheta}(s) \cdot f^q_{\uptheta}(s') \cdot \langle \phi(s), \phi(s') \rangle_{\mathscr{H}} } \nonumber
\end{align}
Applying the kernel trick 
(\ref{defn:kernel_trick}),
\begin{align}
    =& \E{v, v' \sim p}{f^p_{\uptheta}(v) \cdot f^p_{\uptheta}(v') \cdot k(v, v') } \nonumber \\
    & - 2\cdot \E{v \sim p, s \sim q}{f^p_{\uptheta}(v) \cdot f^q_{\uptheta}(s) \cdot k(v, s) } \nonumber \\
    & + \E{s, s' \sim q}{f^q_{\uptheta}(s) \cdot f^q_{\uptheta}(s') \cdot k(s, s') } \nonumber
\end{align}
Our loss of generic summarization $\mathscr{L}^{\mathrm{t}}(V^{\mathrm{t}}, S^{\mathrm{t}}, \uptheta)$ is $\mathrm{MMD}^2_{\mathscr{F}}(V^{\mathrm{t}}, S^{\mathrm{t}}, \uptheta)$.
Recalling $n_{\mathrm{t}} = |V^{\mathrm{t}}|$ and $m_{\mathrm{t}} = |S^{\mathrm{t}}|$:
\begin{align}
    \mathscr{L}^{\mathrm{t}} =& 
    \frac{1}{n_{\mathrm{t}} ^2}\sum_{v, v'}{f_\uptheta^{n_{\mathrm{t}}}(v) \cdot  f_\uptheta^{n_{\mathrm{t}}}(v') \cdot k(v, v') } \nonumber \\
    &-\frac{2}{n_{\mathrm{t}} \cdot m_{\mathrm{t}}}\sum_{v, s}{f_\uptheta^{n_{\mathrm{t}}}(v) \cdot f_\uptheta^{m_{\mathrm{t}}}(s) \cdot k(v, s) } \nonumber \\
    &+\frac{1}{m_{\mathrm{t}}^2}\sum_{s, s'}{f_\uptheta^{ m_{\mathrm{t}} }(s) \cdot  f_\uptheta^{ m_{\mathrm{t}} }(s') \cdot k(s, s') } \nonumber
\end{align}

\section{Training details}
\label{ssec:train_details_appendix}
We train generic summarization model with full batch LBFGS~\citep{liu1989limited} with learning rate $0.005$.
We train comparative summarization model using Yogi optimizer~\citep{NIPS2018_8186}, with a mini batch size of 8 topics, learning rate $0.002$, and
decreasing the learning rate by half every 20 epochs. 
We choose the number of training epochs by validating across 5 folds with early stopping. We set the patience to 20 epochs for early stopping with LBFGS optimizer and 50 epochs with Yogi optimizer. 
We tune the other hyperparameters on the training set, and the optimal hyperparameters of best model (SupMMD + MKL) and searched space are shown in Table~\ref{table:hyperparams}.
The kernel combination weights $\mathbf{w}$ (\S\ref{ssec:mkl}) are also shown in Table~\ref{table:hyperparams}.
The kernel combination weights ($\mathbf{w}$) are written in order: unigrams, bigrams and entities. 
\begin{table}[h]
    \small 
    \centering 
    \begin{tabular}{r|c|c|c|}
    \toprule 
    hyp.         & DUC-2003           & TAC-2008-A       & TAC-2009-B    \\
    \midrule
    $\gamma$     & 2.5[1-4]        & 4.5[2-6]            & 2.2[1-3]   \\
    $\beta$      & 0.04[.02-.16] & 0.08[.02-.16]    & 0.02[.01-.16] \\
    $\lambda$    &    -                &      -          & 0.5[.25-.625] \\
    $r$          & 0.001[0-.01]     & 0.01[-0.01]      & 0.01[-0.01] \\
    epoch        & 64 & 53 & 94 \\
    \midrule
    $\mathbf{w}$ & [.0, .968, .032] & [.01, .97, .02] & [.014, .98, .006]\\
    \bottomrule
    \end{tabular}
    \caption{Optimal hyperparameters, their search space and MKL combination weights on each dataset.}
    \label{table:hyperparams}
\end{table}

\begin{table*}[h]
    \centering
    \small
    \begin{tabular}{r|rr|rr|rr}
        \toprule
        & \multicolumn{2}{c|}{DUC2004} & \multicolumn{2}{c|}{TAC2009-A} & \multicolumn{2}{c}{TAC2009-B} \\
        feature & SupMMD & LexRank &  SupMMD & LexRank &  SupMMD & LexRank\\
        \midrule
        position & 0.34 & 0.16 & 0.32 & 0.18 &  0.44 & 0.22 \\
        tfisf    & 0.07 & 0.38 & 0.22 & 0.37 &  0.01 & 0.36 \\
        btfisf   & 0.30 & 0.52 & 0.48 & 0.53 &  0.46 & 0.57 \\
        \#words  & 0.0  & 0.35 & 0.08 & 0.33 & -0.15 & 0.31 \\ 
        \#nouns  & 0.15 & 0.43 & 0.27 & 0.41 &  0.08 & 0.40 \\             
        \bottomrule
    \end{tabular}
    \caption{Correlation of some features with sentence scores from SupMMD and Lexrank eigenvector centrality. }
    \label{table:feat_corr}
\end{table*}

\section{Additional results}
\label{sec:additional_results}
In this section we provide some additional results.

\begin{table*}[t]
    \small 
    \centering
    \begin{tabular}{r|p{6.75cm}|p{6.75cm}|}
        \toprule 
        \bf{method} & \bf{set A} & \bf{set B}\\
        \midrule 
        ICSI & {\textbf{A fourth day} of thrashing thunderstorms began to take a heavier toll on southern California with at least \textbf{three deaths} blamed on the rain, as flooding and mudslides forced road closures and emergency crews carried out harrowing rescue operations.  Downtown Los Angeles has had more than 15 inches of rain since Jan. 1, more than its average rainfall for an entire year, including 2.6 inches, a record. Meteorologists say Southern California has not been hit by \textbf{this much rain in nearly 40 years}. The disaster was the latest caused by \textbf{rain and snow} that has battered California since Dec. 25.} & {Californians braced for \textbf{even more rain} as they struggled to recover from storms that have left at least \textbf{nine people dead}, triggered mudslides and tornadoes, and washed away roads and runways. The \textbf{record}, 38.18 inches (96.98 centimeters), was set in \textbf{1883-1884}. \textbf{Mudslides} forced Amtrak to suspend train service between Los Angeles and Santa Barbara through at least Thursday. A \textbf{winter storm} pummeled Southern California for the third straight day, claiming the lives of three people and raising fears of mudslides, even as homes around the region were evacuated. Staff Writers Rick Orlov and Lisa Mascaro contributed to this story.}\\
        \midrule
        SupMMD & {Downtown Los Angeles has had more than \textbf{15 inches of rain} since Jan. 1, more than its average rainfall for an entire year, including 2.6 inches, a record. A \textbf{fourth day} of thrashing thunderstorms began to take a heavier toll on southern California with at least three deaths blamed on the rain, as \textbf{flooding and mudslides} forced road closures and emergency crews carried out harrowing rescue operations. The roads in Los Angeles County were equally frustrating. Part of a rain-saturated hillside gave way, sending a Mississippi-like \textbf{torrent of earth and trees} onto four blocks of this oceanfront town and killing two men.} & {Storms have caused \textbf{\$52.5 million} (euro39.8 million) in damage to Los Angeles County roads and facilities since the beginning of the year. Multi-million-dollar homes collapsed and mudslides trapped residents in their homes as a heavy rains that have claimed three lives pelted Los Angeles for the fifth straight day. In scenes reminiscent of the aftermath of the Northridge Earthquake 11 years ago this month, Los Angeles area residents faced gridlocked freeways and roads Wednesday while cleanup crews cleared mud, rubble and debris left from a \textbf{two-week siege of rain}. A \textbf{record-shattering storm} slammed Southern California for a \textbf{sixth straight day} Tuesday, triggering \textbf{mudslides and tornadoes} and forcing more road closures, but forecasters predicted it would wane Wednesday before a new storm moves in Sunday night.}\\
        \bottomrule
    \end{tabular}
    \caption{Example summaries of topic D0906, containing articles about "Rains and mudslides in Southern California".}
    \label{table:examples}
\end{table*}

\subsection{Correlation with rouge score}\label{ssec:res_rouge_corr}
\begin{table}[H]
    \centering
    \small
    \begin{tabular}{r|rr|rr}
        \toprule
        & \multicolumn{2}{c|}{ROUGE-2} & \multicolumn{2}{c}{ROUGE-1} \\
        dataset    & SupMMD & LexRank & SupMMD & LexRank \\
        \midrule
        TAC2009A & 0.590  & 0.555   & 0.571    & 0.543   \\
        DUC2004   & 0.595  & 0.577   & 0.567    & 0.545   \\ 
        \bottomrule
    \end{tabular}
    \caption{Correlation of sentence importance scores with normalized sentence ROUGE scores. }
    \label{table:rouge_corr}
\end{table}
We analyze the correlation between normalized ROUGE recall scores of the sentences and sentence scores from \textit{SupMMD} and \textit{Lexrank}. 
The normalized rouge score of each sentence is defined as $\overline{\rm ROUGE}(s) = \frac{{\rm ROUGE}(s)}{\# {\rm words}(s)}$.
As shown in Table \ref{table:rouge_corr},
we find that \textit{SupMMD} has a slightly high correlation with sentence rouge scores. 
This suggests that \textit{SupMMD} is better in capturing sentence importance for summarization.


\subsection{Feature correlations}\label{ssec:res_pos}
We analyze the correlation between various surface features and sentence importance scores from \textit{SupMMD} and \textit{Lexrank}~\citep{erkan2004lexrank}. 
As shown in table \ref{table:feat_corr}, 
\textit{SupMMD} has higher correlation with relative position, signifying the importance of position of sentence in summary sentences. \textit{Lexrank} has a higher correlations with the number of words, number of nouns and TFISF scores of the sentences, which is expected as \emph{Lexrank} is an eigenvector centrality of sentence-sentence similarity matrix. 
This suggest \emph{SupMMD} is able to learn that first few sentences are important in news summarization.
Similar result is reported by~\citet{kedzie-etal-2018-content}, where they show that the first few sentences are important in creating summary of news articles.



\subsection{Example summary}
\label{ssec:example_appendix}
We present the update summaries (Set A and B) of topic D0906, which contains articles about \textit{"Rains and mudslides in Southern California"} in Table \ref{table:examples}.
We highlight few phrases in bold which could help us to identify the difference between set A and B.
Summaries from \emph{ICSI} and \emph{SupMMD} methods suggest that set A contains articles describing events from earlier days of the disaster and set B contains articles from later stage of the disaster.

\end{document}